\documentclass{article}

\usepackage[final]{neurips_2021_safe_rl}

\usepackage[utf8]{inputenc} 
\usepackage[T1]{fontenc}    
\usepackage{hyperref}       
\usepackage{url}            
\usepackage{booktabs}       
\usepackage{amsfonts}       
\usepackage{nicefrac}       
\usepackage{microtype}      
\usepackage{xcolor}         
\usepackage{graphicx}
\usepackage{amsmath}
\usepackage{amsthm}
\usepackage{amssymb}

\title{Execute Order 66: Targeted Data Poisoning for Reinforcement Learning via Minuscule Perturbations}
\author{Harrison Foley \\
    United States Navy \\
    \texttt{harrison.foley99@gmail.com}
    \And 
    Liam Fowl \\
    Department of Mathematics \\
    University of Maryland \\ 
    \texttt{lfowl@umd.edu}
    \And
    Tom Goldstein \\
    Department of Computer Science \\
    University of Maryland \\
    \texttt{tomg@cs.umd.edu}
    \And
    Gavin Taylor \\
    Department of Computer Science \\
    United States Naval Academy \\
    \texttt{taylor@usna.edu}
}

\begin{document}

\maketitle

{\bf  Abstract}

Data poisoning for reinforcement learning has historically focused on general performance degradation, and targeted attacks have been successful via perturbations that involve control of the victim's policy and rewards. We introduce an insidious poisoning attack for reinforcement learning which causes agent misbehavior only at specific target states - all while minimally modifying a small fraction of training observations without assuming any control over policy or reward. We accomplish this by adapting a recent technique, gradient alignment, to reinforcement learning. We test our method and demonstrate success in two Atari games of varying difficulty. 

\section{Introduction}\label{sec:intro}

Reinforcement Learning (RL) has quickly achieved impressive results in a wide variety of control problems, from video games to more real-world applications like autonomous driving and cyber-defense \citep{alphastarblog, galias2019simulation, DBLP:journals/corr/abs-1906-05799}. However, as RL becomes integrated into more high risk application areas, security vulnerabilities become more pressing. 
 
One such security risk is data poisoning, wherein an attacker maliciously modifies training data to achieve certain adversarial goals. In this work, we carry out a novel data poisoning attack for RL agents, which involves imperceptibly altering a small amount of training data. The effect is the trained agent performs its task normally until it encounters a particular state chosen by the attacker, where it misbehaves catastrophically. Although the complex mechanics of RL have historically made data poisoning for RL challenging, we successfully apply gradient alignment, an approach from supervised learning, to RL \citep{geiping2020witches}. Specifically, we attack RL agents playing Atari games, and demonstrate that we can produce agents that effectively play the game, until shown a particular cue. We demonstrate that effective cues include a specific target state of the attacker's choosing, or, more subtly, a translucent watermark appearing on a portion of any state. 

\section{Reinforcement Learning}\label{sec:rl}

In reinforcement learning, the reinforcement learner, or agent, makes decisions based on the situation it is in and its own understanding of which actions will lead to positive outcomes in that situation in order to achieve some task \citep{sutton2018reinforcement}. 

Formally, we model RL as a Markov Decision Process (MDP). We describe this by the 5-tuple $M = (S, A, P, R, \gamma)$. For a specific scenario, $S$ represents the set of all possible states, and $A$ represents the set of all possible actions. Given states $s_i,s_j \in S$, $P(s_j|s_i,a)$ gives the probability of a transition to $s_j$ from $s_i$ after taking an action $a \in A$. $R(s_i)$ represents the expected reward associated with state $s_i$; for example, a state in a checkers game that represents a win would receive a reward of 1, a state that represents a loss would receive a reward of -1, and all other states would receive a reward of 0. Finally, $\gamma$, known as the discount factor, is a constant between 0 and 1 that describes how significant future rewards are compared to immediate rewards. 

By exploring this MDP, the agent must learn behavior which will result in accomplishing the task or achieving good results. This behavior is defined by the policy, $\pi(s)$, which maps each state $s\in S$ to an action $a\in A$, describing which action the agent will take in a given state. Policies may also be stochastic, in that they map to a distribution over actions. In this case, the action chosen would be sampled from this distribution. We can evaluate or improve a policy using the value function $V_\pi(s)$. The value function for a policy $\pi$ is given by the Bellman equation,
\begin{equation}\label{eqn:Bellman}
V_\pi(s_i) = R(s_i) + \gamma \sum_{s_j \in S}P(s_j|s_i,a\sim\pi(s_i))V_\pi(s_j).
\end{equation}

The value function depends on $\pi(s)$; the goal of the agent is to build a policy that results in high value. A common approach to training RL agents is to have an agent explore it's environment. During learning, the agent generates samples of the form $(s,a,r,s')$, where $s$ is  the state the agent begins in, $a$ is an action drawn from the distribution given by the policy at state $s$ ($a\sim\pi(s)$), $r$ is the reward received, and $s'$ is the state the agent ends up at as a result of the action.  Using a large number of these samples, the agent simultaneously attempts to make its approximate value function for the policy more accurate, and improves the policy to take actions the value function indicates are more valuable. A common approach for building a good policy is to use a policy gradient scheme in which a neural network parameterizes and approximates the policy. The policy is then optimized via gradient descent on an objective function.

\section{Data Poisoning: Prior Work}\label{sec:dp}

Data poisoning attacks on machine learning models involves attackers modifying training data in order to achieve malicious goals. In the setting of supervised learning (SL), data poisoning has been extensively studied. Here, data poisoning can be roughly categorized into \emph{availability} attacks and \emph{integrity} attacks \citep{barreno_security_2010}. Availability attacks aim to degrade the general performance of a machine learning model.  Early availability attacks focused on simple settings like logistic regression, and support vector machines \citep{biggio_poisoning_2012, munoz-gonzalez_towards_2017}. Recently, heuristics have been leveraged to perform availability attacks on deep networks \citep{feng2019learning, huang2021unlearnable, fowl2021preventing}. In contrast to availability attacks, integrity attacks focus on causing a victim model to misclassify a \emph{select} set of targets. This attack can be more insidious than an availability attack since there is no noticeable drop in performance of the model trained on the poisoned data, and the effects of the poisons are only realized when a specific target is seen by the model \citep{geiping2020witches}. 

Data Poisoning for RL is more challenging because of a seemingly more fragile trade-off between the success of the attack and the overall abilities of the model. Perhaps because of this trade-off, many previous attacks have focused on availability, or preventing the agent from learning any good behavior at all \citep{sun2021vulnerabilityaware}. Given the various challenges and instability in training neural networks for RL \citep{Ding2020}, over-modifying training data could have catastrophic effects to the learning process, which is not desirable if the attacker seeks to preserve good behavior of the agent except for a specific condition that the attack is designed to be effective for.

Additionally, in a policy gradient scheme, it is standard to discard older samples which no longer align with the current policy as the agent trains. As a result, poisoned data generated under obsolete policies can no longer influence training. Ultimately, this means that in order to carry out a successful an attack for an agent training via a policy gradient scheme, new poisons must be generated online for the current parameterization of the policy.

There have been successful integrity attacks for reinforcement learning, but these attacks have required alterations to the states encountered, actions chosen, and rewards received by agents during training \citep{kiourti2019trojdrl}. In other words, the attacked data looked noticeably different than clean data to a potential observer, and the attacks assumed omnipotent control of the policy and reward.

\section{Our Approach}

In contrast to previous work, we leverage advancements in the data poisoning literature to carry out targeted attacks via minimal perturbations to a select few states encountered by an agent without any direct manipulation of the policy or reward functions. Recently, \emph{gradient alignment} has been adopted as a technique to solve certain instances of bi-level optimization problems in machine learning, such as the creation of synthetic data, and poisoning for supervised learning \citep{zhao2020dataset, geiping2020witches}. Gradient alignment involves perturbing input data to a machine learning model in order to align the training gradient with the gradient of some target objective (the target gradient) w.r.t. the model parameters. We leverage the mechanics of gradient alignment in the context of RL. Simply put, under certain conditions (see \citet{geiping2020witches}) if a network's minibatch gradient aligns with the target gradient, then training on that minibatch of perturbed data also decreases the target objective.

\subsection{Gradient Alignment Poisoning for RL}

For an RL agent, we denote the training loss function, $\mathcal{L}_{train}(\mathcal{D}, \theta)$, where $\mathcal{D}$ is the training data observed by the network and $\theta$ represents the current parameters of the network. 

Conversely, the \emph{adversarial loss}, $\mathcal{L}_{adv}(s_{t}, \theta, a_{d})$ measures whether the agent takes the desired misbehavior $a_d$ at the predetermined target state, $s_{t}$. How the attacker implements this function is can vary. 

The goal of our attack is to perturb existing training data such that the resulting data causes alignment between the gradients of $\mathcal{L}_{train}$ and $\mathcal{L}_{adv}$, so that when the agent decreases $\mathcal{L}_{train}$ through gradient descent, it simultaneously and unknowingly decreases $\mathcal{L}_{adv}$.

To do this, an attacker needs to optimize a third loss function called the alignment loss, $\mathcal{L}_{align}(\nabla_{\theta}\mathcal{L}_{train}(\mathcal{D}, \theta), \nabla_{\theta}\mathcal{L}_{adv}(s_{t}, \theta, a_{d}))$, which quantifies how well the gradients are aligned. Specifically, we choose to optimize the cosine similarity between the two gradient vectors: 

\begin{equation}\label{eqn:NegCosSim}
  \mathcal{L}_{align} = 1 - \dfrac{\nabla_{\theta}\mathcal{L}_{train}(\mathcal{D}, \theta) \cdot \nabla_{\theta}\mathcal{L}_{adv}(s_{t}, \theta, a_{d})}{|\nabla_{\theta}\mathcal{L}_{train}(\mathcal{D}, \theta)||\nabla_{\theta}\mathcal{L}_{adv}(s_{t}, \theta, a_{d})|},
\end{equation}

However, in order to enforce that any perturbations the attacker makes are not detectable to an observer, we also impose constraints on the attacker's perturbations. Usually, these constraints are on the amount of data that is able to be perturbed, and on the $\ell_\infty$ norm of the perturbation. Thus, the attacker aims to optimize the following objective: 

\begin{gather*}\label{eqn:DPopt}
  \min_{\Delta\mathcal{D}} \mathcal{L}_{align}\left(\nabla_{\theta}\mathcal{L}_{train}(\mathcal{D}+\Delta\mathcal{D}, \theta), \nabla_{\theta}\mathcal{L}_{adv}(s_{t}, \theta, a_{d})\right),   \\
  s.t.\ ||\Delta\mathcal{D}||_{\infty} < \epsilon
\end{gather*}
where $\Delta\mathcal{D}$ is the change in the training data, and $\epsilon$ is a small number representing the maximum amount each individual pixel of an observation is allowed to change.

Additionally, we enforce that only a small fraction of the data is allowed to be perturbed. We specify these constraints for individual experiments. We optimize the attacker's objective using projected gradient descent (PGD).

\subsection{Attack Scenarios and Implementation Details}\label{sec:imp}

We assume the attacker has access to the victim agent's parameters but cannot control the victim network directly. The attacker can only modify (minimally) a small portion of observations. We also assume that an observer is present and we therefore minimize the visibility of our attacks via small perturbations. While we do assume white-box access to the agent's parameters, we present a poisoning framework that is more realistic than control over the agent's reward and policy.

For environments involving image input states - as is the case for Atari games - we constrain the observation modifications to the training data in $\ell_\infty$ norm, as is common in supervised learning vision threat models \citep{madry2017towards}.

We present two example attacks.  In the first, we choose a specific state for the agent to misbehave in, which is not encountered during training. As an analogy, imagine an RL agent being trained to drive a self-driving car.  The attacked state might be a scene from a particular road which is unlikely to be experienced during training; successfully attacked agents would drive cars capably everywhere except that particular road, where it would behave catastrophically.

In the second attack, we attempt an attack in which the vulnerability is instead to a particular feature included on any state. In this case, the analogy is the attacked autonomous vehicle driving capably until it encounters a specific billboard that has been erected on any road. In our experiments, we use a specific watermark appearing over the victim agent's observation.

For our RL training algorithm, we chose the Proximal Policy Optimization algorithm \citep{schulman2017proximal} as implemented by stable-baselines \citep{hill2018stable}; PPO has demonstrated state-of-the-art (and frequently superhuman) performance on Atari games. For $\mathcal{L}_{adv}$, we use a softmax cross-entropy function between the action probability distribution at the target state as output by the policy network and the action probability distribution we wanted to induce. This is because PPO's policies are stochastic, meaning that they return an action probability distribution that the agent samples from rather than a single action. In our case, when the attacker aims to induce a particular action, we compare the agent's action probability distribution to a distribution where the probability of $a_{d}$ was 100\% and all other actions had probability 0\%. Finally, we optimize $\mathcal{L}_{align}$ via signed PGD for $\epsilon$ iterations, storing the poisoned training observations that yield the lowest value for $\mathcal{L}_{align}$, and using those observations as our final poisoned training observations.

\begin{figure}[!h]
\minipage{0.32\textwidth}
  \includegraphics[width=\linewidth]{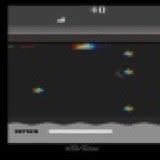}
\endminipage\hfill
\minipage{0.32\textwidth}
  \includegraphics[width=\linewidth]{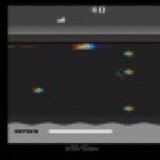}
\endminipage\hfill
\minipage{0.32\textwidth}%
  \includegraphics[width=\linewidth]{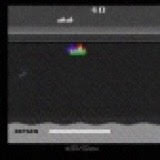}
\endminipage
\caption{Illustration of perturbations used for Seaquest: clean (leftmost), $\epsilon = 1$ (center), meaning each color value in each pixel are allowed to increase or decrease by at most 1, and $\epsilon=8$ (rightmost), meaning each color value of each pixel is allowed to increase or decrease by at most 8. Color values for each pixel range from 0 to 255, so these perturbations are very hard to notice.}
\label{perturb_seaquest}
\end{figure}

\section{Experimental Results}

Here, we outline the different attack scenarios and domains we study. 

\subsection{First Attack Scenario: Arbitrary Target State}

\begin{figure}[h!]
\begin{center}
  \includegraphics[width=5in]{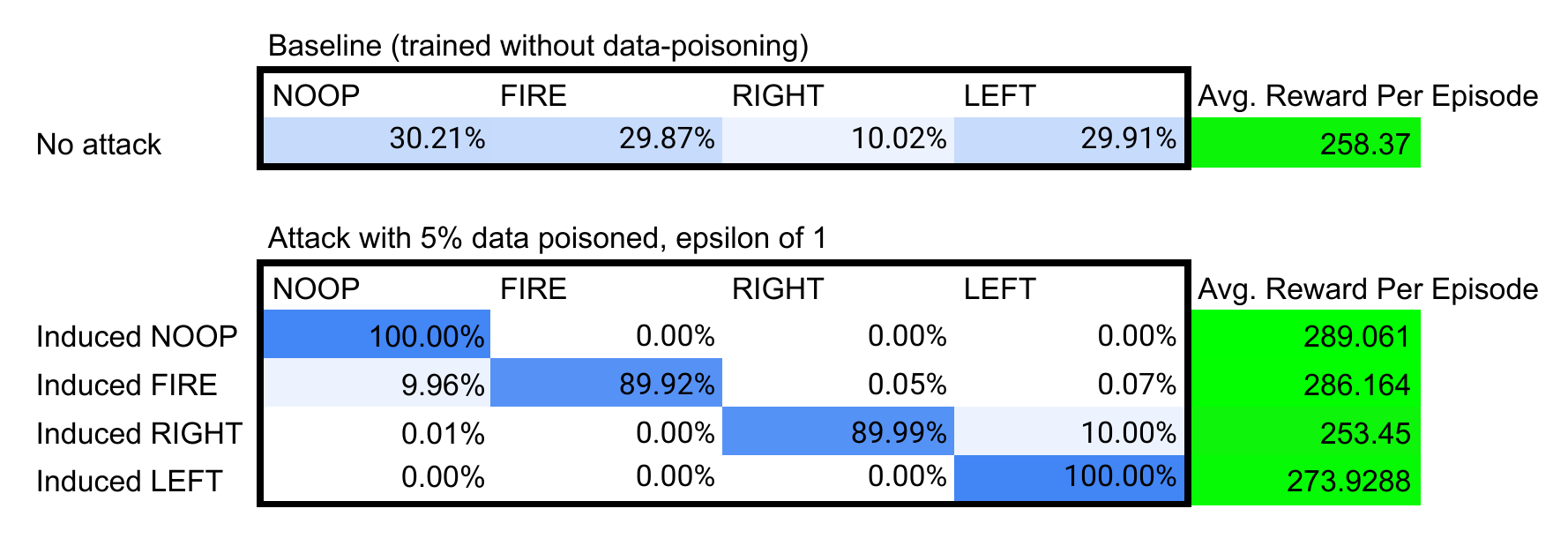}
\end{center}
  \caption{Breakout results - \textbf{top:} network trained from scratch with no poisoning, \textbf{bottom:} results for $\epsilon = 1/255$ attack. Each row is averaged over 10 models trained and poisoned in order to induce the same action (except for the top table). The portions heatmapped in blue represent the average action probability distribution at a the target state.}
\label{fig:DPresBreakout}
\end{figure}

In our first scenario, we choose an arbitrary state in the future for $s_{t}$. The first domain we carried out this experiment for was the Atari game ``Breakout," in which the agent is trained to control a paddle at the bottom of the screen to rebound a ball into bricks at the top of the screen. We began with this domain because it is simple; the states are not varied, and the agent only can choose 1 of 4 actions at a given moment. The agent can move the paddle left, move the paddle right, leave the paddle where it is, or start the game \citep{Bellemare_2013}. As shown in Figure \ref{fig:DPresBreakout}, the state chosen for $s_{t}$ appears to be inconsequential given the fact that a variety of well-trained agents pick all 4 actions fairly evenly. This suggests that a successful data poisoning attack would be easier to achieve, in theory.  We achieved this success as shown in Figure \ref{fig:DPresBreakout}, for even with a very small attack, we were able to decisively induce all 4 actions without decreasing the overall performance of the agents.

\begin{figure}[!h]
\minipage{0.32\textwidth}
  \includegraphics[width=\linewidth]{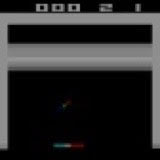}
\endminipage\hfill
\minipage{0.32\textwidth}
  \includegraphics[width=\linewidth]{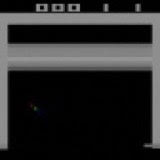}
\endminipage\hfill
\minipage{0.32\textwidth}%
  \includegraphics[width=\linewidth]{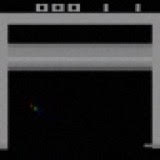}
\endminipage
\caption{Illustration of perturbations used for Breakout: clean (leftmost), $\epsilon = 1$ (center), meaning each color value in each pixel are allowed to increase or decrease by 1, and $\epsilon=8$ (rightmost), meaning each color value of each pixel is allowed to increase or decrease by at most 8. Color values for each pixel range from 0 to 255, so perturbations are very hard to notice.}
\label{perturb_breakout}
\end{figure}

\begin{figure}[h!]
\begin{center}
  \includegraphics[width=5in]{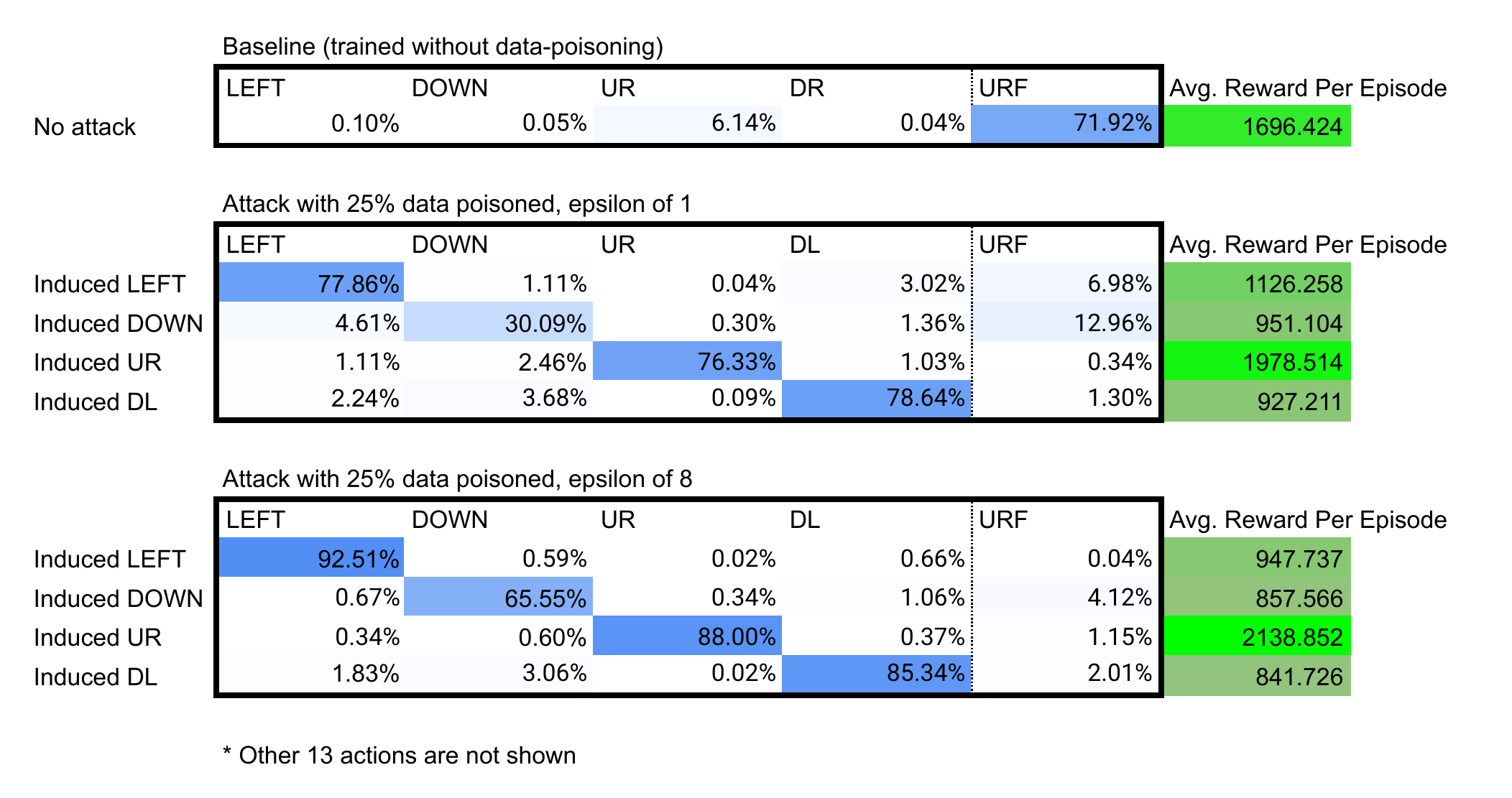}
\end{center}
  \caption{Seaquest results - \textbf{top:} network trained from scratch with no poisoning, \textbf{middle:} results for $\epsilon=1/255$ attack, \textbf{bottom:} results for $\epsilon=8/255$ attack. Each row is averaged over 10 models trained and poisoned in order to induce the same action (except for the top table).}
\label{fig:DPresSeaquest}
\end{figure}

The second domain we considered was the Atari game ``Seaquest". There are 18 actions the agent can choose from at any given moment. This domain is far more difficult than Breakout, as the agent must learn more complex behavior in order to thrive. As indicated in Figure \ref{fig:DPresSeaquest}, the state chosen for $s_{t}$ appears to be more significant/consequential than the one chosen for Atari; several well-trained agents all picked the 'up-right-fire' action, which suggests that this action choice at this moment is important for successful gameplay. This suggests that a data poisoning attack for this state would be more challenging. Additionally, in order to present a greater challenge to the attacker, we attempted to induce some behavior that was very different from 'up-right-fire', such as left, down, and down-left.

We find that the attack size needed to be larger in most cases in order to induce our desired behavior (see Figure \ref{fig:DPresSeaquest}). This more potent attack in turn yielded agents that performed noticeably worse at the game than agents trained without poisoning, with the exception of those poisoned to induce up-right, possibly because the difference in choice between up-right and up-right-fire as actions is relatively inconsequential. However, with the more potent attack, we find that poisoning in this scenario was largely successful in that we induced the desired misbehavior decisively at each target state of each domain.

\subsection{Second Attack Scenario: Target Watermark}

In addition to poisoning for an arbitrary target state, $s_{t}$, we carried out attacks that aimed to induce behavior for a particular watermark (we chose a watermark of straight vertical lines) appearing over-top of part of \textit{any} state. This is similar to a \emph{backdoor} attack for supervised learning wherein a ``trigger" is added to target images in order to cause misclassification \citep{turner2018clean}. The watermark covers 25\% of the area of any state. While this scenario would not happen organically in games, it is analogous to a self-driving car's camera picking up a particular billboard on the side of the road. 

\begin{figure}[ht]
\begin{center}
  \includegraphics[width=5in]{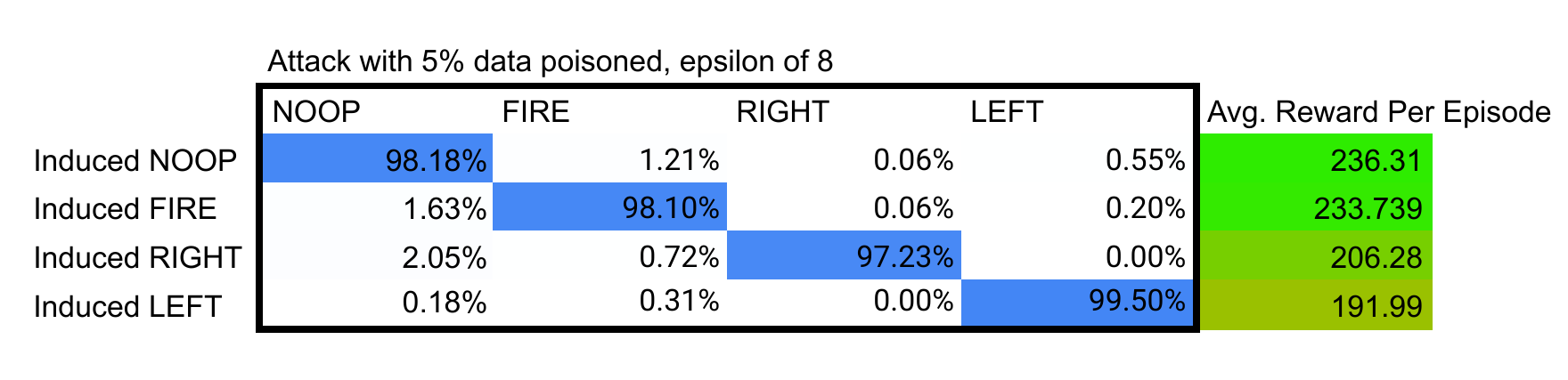}
\end{center}
  \caption{Breakout results for watermark scenario - Each row is averaged over 4 models trained and poisoned in order to induce the same action. The portions heatmapped in blue represent the average action probability distribution at a selection of 1024 arbitrary states with the watermark present for the 4 networks in each row.}
\label{fig:Breakout_bb}
\end{figure}

\begin{figure}[ht]
\begin{center}
  \includegraphics[width=5in]{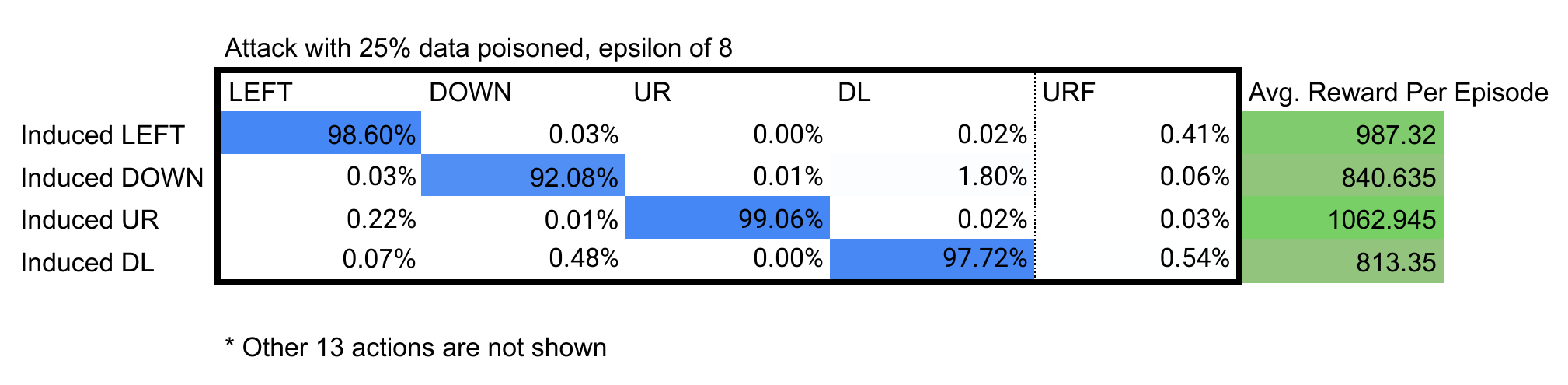}
\end{center}
  \caption{Seaquest results for watermark scenario - Each row is averaged over 4 models.}
\label{fig:Seaquest_bb}
\end{figure}

This attack was also largely successful for Breakout. Figure \ref{fig:Breakout_bb} shows that the watermark decisively induced the desired behavior and only caused minor performance drops. The attack was also successful for Seaquest as shown in Figure \ref{fig:Seaquest_bb}, though the larger attack used did cause performance drops in comparison to agents trained without any data poisoning at all (see the top tables of Figures \ref{fig:DPresBreakout} and \ref{fig:DPresSeaquest} for baseline scores). The magnitude of this performance impact depended on the similarity of the action to the up-right-fire preferred by the clean agent.  The greater the difference between up-right-fire and the induced action, the greater the performance impact.  All agents, however, remained capable of playing the game without any watermarks shown with reasonable success.

\section{Ongoing Work}\label{sec:ow}

\begin{figure}[h]
\begin{center}
  \includegraphics[width=5in]{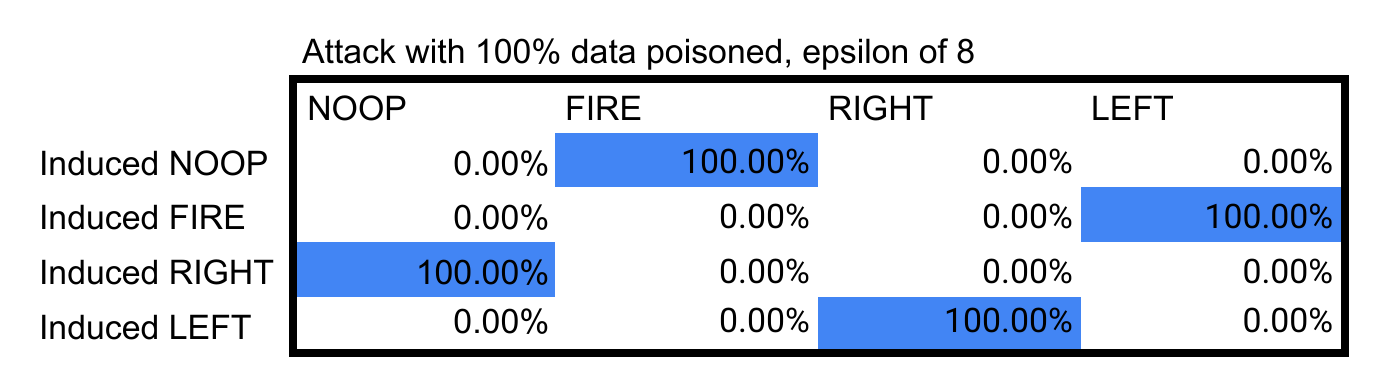}
\end{center}
  \caption{A failed black-box attacked for an agent trained to play Breakout. All of the training data was poisoned with $\epsilon = 8$, and yet the attack was still unsuccessful at inducing the desired misbehavior in any case.}
\label{fig:black-box-fail}
\end{figure}

In order to continue this research, a first priority would be to successfully carry out poisoning in a black-box scenario, in which we do not have access to the agent itself and so cannot compute gradients or fashion poisons for it directly. Our current approach requires that the attacker has direct access to the victim agent. This is practically more challenging than a black-box attack, which would require us to fashion poisons for a known agent or ensemble of agents, in such a way that the poisoned data would also be effective on a separate, unknown attacked agent. Poisons created in our initial experiments in this setting have not successfully transferred to a black-box agent; Figure \ref{fig:black-box-fail} shows the that in each case, the agent does not choose our desired misbehavior in Breakout  when the gradients used to compute an attack are done so on a separate black-box agent. 

While it is difficult to speculate why these attacks are unsuccessful, a reasonable hypothesis is that the black-box agent and white-box victim agent aren't perfectly synchronized in their respective training stages. In addition, the black-box agent is likely randomly initialized differently than the white-box agent, which could also contribute to the fact that a successful attack on the black-box agent are not effective on our victim agent. In order to combat this, a black-box attack might need to be generated by an ensemble of many different black-box agents at once in order for the attack to work for any agent, not just a specific agent. 

Second, further work could include attempting to replicate this effect in even more Atari games or other domains, as well as various learning algorithms, to confirm that the vulnerability exists in various applications of RL. Specifically, a next step we could include carrying out an attack for an agent in MuJoCo, a physics engine designed for RL agents to learn to walk/move/accomplish tasks in a 3D space. Our attacks might include inducing the agent to fall over at a particular state, but to walk normally otherwise. This would demonstrate the ability to craft poisons capable of hiding in a lower-dimensional state space than images of Atari games.

\section{Conclusion}

This work raises questions about the degree to which deep neural networks deployed for reinforcement learning are secure. Successful online attacks for the two tested domains required only minuscule perturbations to a very small fraction of the training data in order to be successful. Additionally, the perturbations only consisted of virtually invisible modifications to the states themselves without any alterations of the policy or reward function, meaning an onlooker would have a very difficult time detecting that any attack took place whatsoever. The implications of a successful poisoning attack in a safety-critical agent, such as an autonomous vehicle, are potentially life-threatening. Therefore, more time must be invested in making neural networks for RL robust to these kinds of attacks before we can reasonably trust RL agents with human lives.

\paragraph{Acknowledgements}

This work was supported by the Office of Naval Research, document numbers
N0001422WX01228 and N000142112557.

\bibliographystyle{plainnat}
\bibliography{main}

\end{document}